\documentclass[sn-standardnature]{sn-jnl}

\jyear{2023}%

\unnumbered

\usepackage{caption}
\usepackage{subcaption}
\usepackage{float} 
\usepackage{makecell}

\usepackage{anyfontsize} 

\newcommand{\proposed}{{AutoPrognosis-M}}

\begin{document}
 
\title[AutoPrognosis-Multimodal]{Automated Ensemble Multimodal Machine Learning for Healthcare}

\author*[1]{\fnm{Fergus} \sur{Imrie}}\email{fergus.imrie@stats.ox.ac.uk}
\equalcont{These authors contributed equally to this work.}
\author[2,3]{\fnm{Stefan} \sur{Denner}}\email{stefan.denner@dkfz-heidelberg.de}
\equalcont{These authors contributed equally to this work.}
\author[4]{\fnm{Lucas S.} \sur{Brunschwig}}\email{lucas.brunschwig@epfl.ch}
\author[2,5,6]{\fnm{Klaus} \sur{Maier-Hein}}\email{k.maier-hein@dkfz-heidelberg.de}
\author[7]{\fnm{Mihaela} \sur{van der Schaar}}\email{mv472@cam.ac.uk}

\affil[1]{\orgname{Department of Statistics, University of Oxford}, \orgaddress{\country{United Kingdom}}}
\affil[2]{\orgname{Division of Medical Image Computing, German Cancer Research Center (DKFZ)}, \orgaddress{\country{Germany}}}
\affil[3]{\orgname{Medical Faculty Heidelberg, Heidelberg University}, \orgaddress{\country{Germany}}}
\affil[4]{\orgname{École Polytechnique Fédérale de Lausanne}, \orgaddress{\country{Switzerland}}}
\affil[5]{\orgname{Pattern Analysis and Learning Group, Department of Radiation Oncology, Heidelberg University Hospital}, \orgaddress{\country{Germany}}}
\affil[6]{\orgname{National Center for Tumor Diseases (NCT) Heidelberg}, \orgaddress{\country{Germany}}}
\affil[7]{\orgname{Department of Applied Mathematics and Theoretical Physics, University of Cambridge}, \orgaddress{\country{United Kingdom}}}

\abstract{
The application of machine learning in medicine and healthcare has led to the creation of numerous diagnostic and prognostic models. However, despite their success, current approaches generally issue predictions using data from a single modality. This stands in stark contrast with clinician decision-making which employs diverse information from multiple sources. While several multimodal machine learning approaches exist, significant challenges in developing multimodal systems remain that are hindering clinical adoption. In this paper, we introduce a multimodal framework, {\proposed}, that enables the integration of structured clinical (tabular) data and medical imaging using automated machine learning. {\proposed} incorporates 17 imaging models, including convolutional neural networks and vision transformers, and three distinct multimodal fusion strategies. In an illustrative application using a multimodal skin lesion dataset, we highlight the importance of multimodal machine learning and the power of combining multiple fusion strategies using ensemble learning. We have open-sourced our framework as a tool for the community and hope it will accelerate the uptake of multimodal machine learning in healthcare and spur further innovation.
\footnotetext{\emph{Preprint}}
}


\maketitle

\section{Introduction}

Medical and healthcare data is increasingly diverse in origin and nature, encompassing patient records and imaging to genetic information and real-time biometrics. 
Machine learning (ML) can learn complex relationships from data to construct powerful predictive models. 
As a result, ML is increasingly being proposed in medicine and healthcare, particularly for diagnostic and prognostic modeling \cite{Abramoff2018,Rajpurkar2022}. 
However, such approaches typically make predictions based on only one type of data \cite{Kline2022} and thus cannot incorporate all available information or consider the broader clinical context.

In contrast, clinicians make decisions based on the synthesis of information from multiple sources, including imaging, structured clinical or laboratory data, and clinical notes \cite{Huang2020}. 
This can be critical for accurate diagnoses and prognoses, and the absence of such information has been shown to result in lower performance and decreased clinical utility in numerous studies \cite{Leslie2000,Castillo2021}. 
While true across healthcare, this is perhaps particularly the case in medical imaging. 
For example, almost 90\% of radiologists reported that additional clinical information was important and could change diagnoses compared to using the imaging alone \cite{Boonn2009}.
Numerous other examples of the importance of clinical context for medical image analysis exist across specialties such as ophthalmology \cite{Wang2018}, pathology \cite{Ombrello2014}, and dermatology \cite{Bergenmar2002}.

Multimodal machine learning integrates multiple types and sources of data, offering a more holistic approach to model development that mirrors clinical decision-making processes.
As a result, while multimodal ML remains in its infancy, models that incorporate multiple data modalities have been developed for several medical domains including cardiology \cite{Li2021}, dermatology \cite{Liu2020}, oncology \cite{Yala2019,Kyono2020}, and radiology \cite{Wu2021}. 
However, technical challenges in developing, understanding, and deploying multimodal ML systems are currently preventing broad adoption in medicine beyond bespoke individual examples.
Thus, techniques that reduce these obstacles could have substantial benefits across numerous applications and clinical use cases.

\begin{figure*}[th!]
  \centering
  \includegraphics[width=0.99\textwidth]{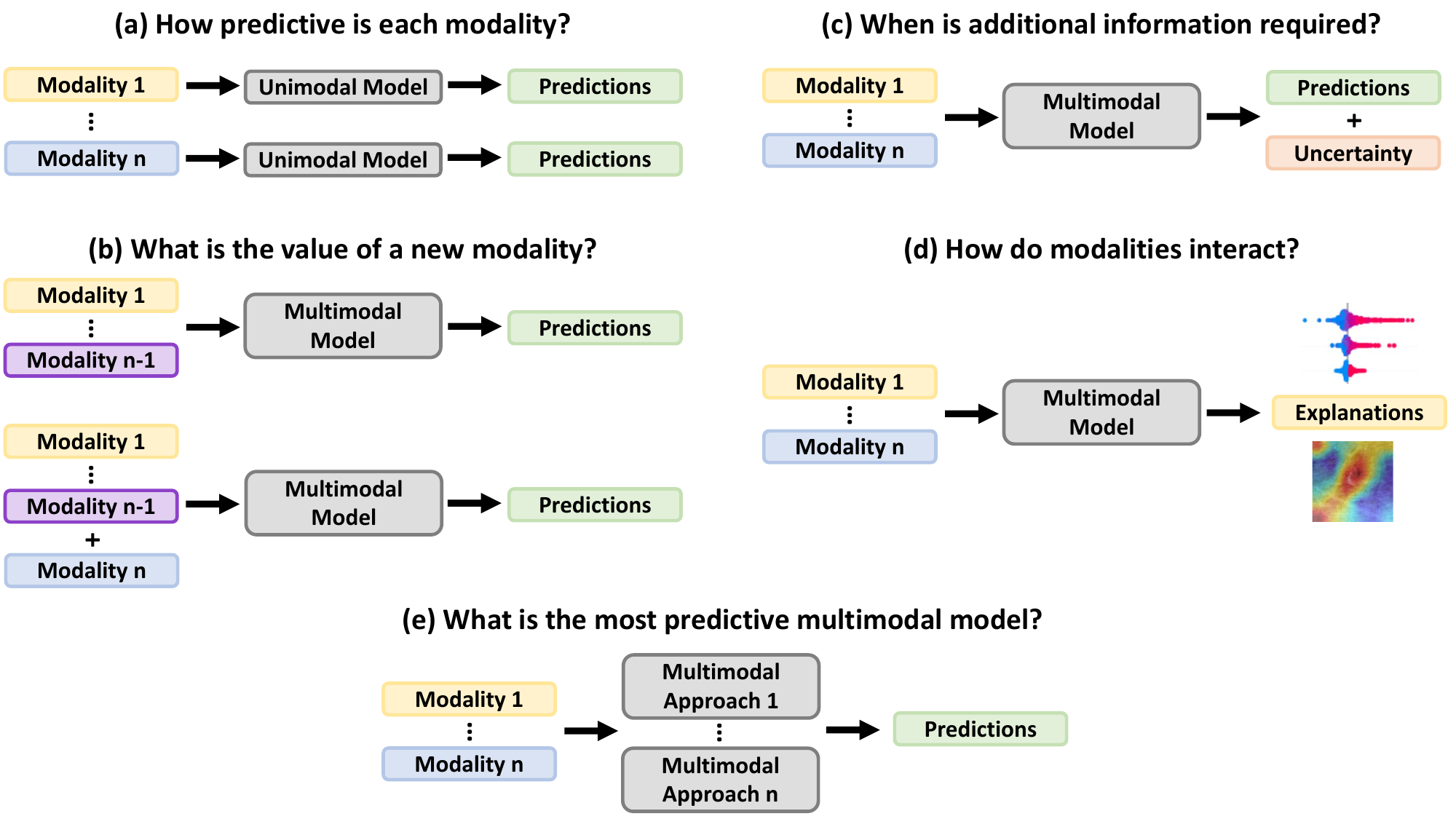} 
  \vspace{0.1cm}
  \caption{\textbf{Overview of the types of questions that can be asked with multimodal machine learning.} In addition to developing powerful multimodal models (e), multimodal ML can help understand the value of each modality (a), the impact of adding a new modality (b), when an additional modality is required (c), and how the information in different modalities interacts (d).}
  \label{fig:multimodal_questions}
  \vspace{0.1cm}
\end{figure*} 

In this paper, we propose a general-purpose approach that addresses these challenges using automated machine learning (AutoML) and ensemble learning.
AutoML can help design powerful ML pipelines by determining the most appropriate modeling and hyperparameter choices, while requiring minimal technical expertise from the user.
To bridge the gap between clinicians and cutting-edge ML, we previously proposed an AutoML approach, AutoPrognosis \cite{Imrie2023AutoPrognosis}, for constructing ensemble ML-based diagnostic and prognostic models using structured data. 
AutoPrognosis has been used to develop clinical models for a number of outcomes, including cardiovascular disease \cite{Alaa2019,imrie2023redefining}, cystic fibrosis \cite{Alaa2018CF}, breast cancer \cite{Alaa2021}, and lung cancer \cite{callender2023assessing}. 
While AutoPrognosis has been shown to yield promising results across a range of medical areas, it is constrained to only handling tabular features. Several other frameworks for automated pipeline optimization, such as Auto-sklearn \cite{feurer2015}, Auto-Weka \cite{Thornton2013}, and TPOT \cite{olson2016tpot}, suffer from the same limitation.

Consequently, in this work, we developed AutoPrognosis-Multimodal ({\proposed}), an AutoML framework that incorporates data from multiple modalities, namely imaging and tabular data.
Additionally, {\proposed} enables such models to be interrogated with explainable AI and provides uncertainty estimates using conformal prediction, aiding understanding and helping build model trust \cite{Imrie2023}.

We applied our approach in an illustrative clinical scenario: skin lesion diagnosis using both images and clinical features.
Our experiments demonstrate the benefit of incorporating information from multiple modalities and highlight the impact of the multimodal learning strategy on model performance.
We show different strategies can be effectively combined to form ensemble models that substantially outperform any individual approach.
Additionally, we quantify the value of information from each modality and show how our framework can be used to determine whether additional data is necessary on an individual patient basis.

While our experiments focus on skin lesion diagnosis, we emphasize that {\proposed} is a general-purpose approach that can be used to train multimodal diagnostic and prognostic models for any disease or clinical outcome, without requiring substantial ML expertise, and can help answer a range of clinical questions (Fig. \ref{fig:multimodal_questions}). 
We have open-sourced {\proposed} as a tool for the community to aid the clinical adoption of multimodal ML models.

\begin{figure*}[th!]
  \centering
  \includegraphics[width=0.99\textwidth, trim={0em 2em 4em 3em}, clip]{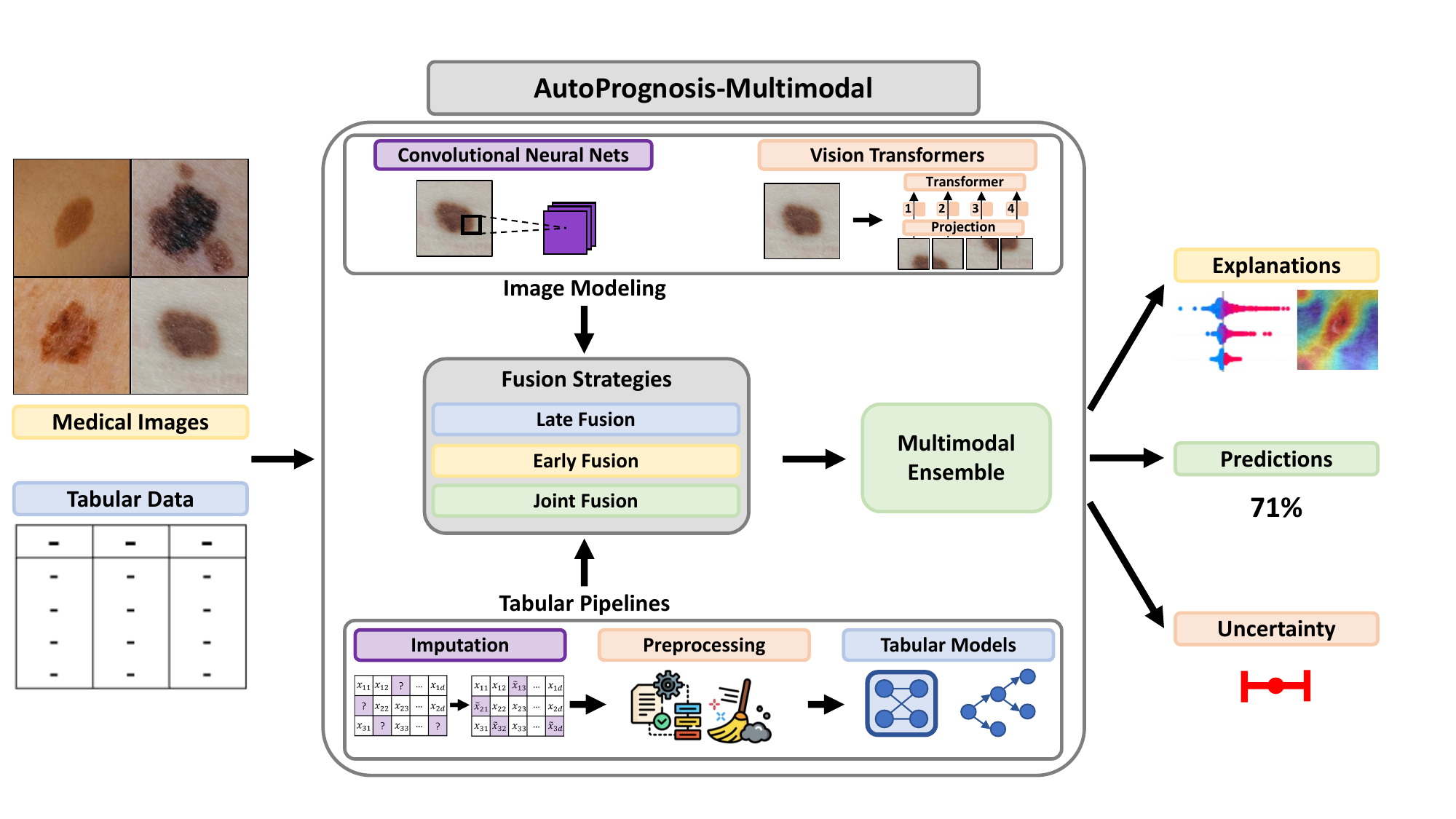} 
  \vspace{0.1cm}
  \caption{\textbf{Overview of {\proposed}.} {\proposed} leverages automated machine learning to produce multimodal ensembles by optimizing state-of-the-art image and tabular modeling approaches across three fusion strategies. {\proposed} also enables such models to be interrogated with explainable AI and provides uncertainty estimates using conformal prediction.}
  \label{fig:autoprognosis-m}
  \vspace{0.1cm}
\end{figure*} 

\section{Methods: {\proposed}}

{\proposed} enables clinicians and other users to develop diagnostic and prognostic models for a diverse range of applications using state-of-the-art multimodal ML (Fig. \ref{fig:autoprognosis-m}).
Perhaps the most significant challenge is the complex design space of model architectures and associated hyperparameters, which must be set appropriately for the specific task and data being considered. Failure to do so can significantly degrade performance; however, this often requires significant ML knowledge and expertise to do so effectively.
This challenge is further compounded in the multimodal setting by the different possible ways of integrating data from multiple sources.

To address this, our framework employs AutoML \cite{Thornton2013} to efficiently and effectively search the model and hyperparameter space.
{\proposed} constructs powerful ensemble multimodal ML models designed for the specific problem and data under consideration.
While incorporating multiple modalities can improve predictive power, this may not always be the case or a modality might be particularly expensive to acquire. Thus, it is important to understand the importance of each modality individually, and the added value of including an additional modality (Fig. \ref{fig:multimodal_questions}a-b). 
{\proposed} can optimize both unimodal models and multimodal models, allowing the value of each modality to be assessed, both separately and in combination with other sources of information, as well as enabling models to be debugged and understood using techniques from explainable AI (Fig. \ref{fig:autoprognosis-m}).

By automating the optimization of ML pipelines across multiple modalities and determining the most suitable way of combining the information from distinct sources, we reduce the barrier for non-ML experts, such as clinicians and healthcare professionals, to build powerful multimodal ML models for problems in the healthcare and medicine domains. 
We believe that {\proposed} significantly simplifies the process of training and validating multimodal ML models without compromising on the expressiveness or quality of the ML models considered.

\subsection{Automated Machine Learning}

AutoML aims to simplify and automate the process of designing and training ML models, thereby reducing the technical capabilities required to develop effective models.
Human practitioners have biases about what model architectures or hyperparameters will provide the best results for a specific task. While this might be helpful in some cases, often it will not and can cause inconsistency in the quality of the final predictive system \cite{Callender2023}. 
AutoML helps minimize these biases by automatically searching over a more general set of models, hyperparameters, and other design choices to optimize a given objective function, returning the best configurations found.
Beyond simply minimizing human biases, AutoML reduces the demand for human experts and has been shown to typically match or exceed the skilled human performance \cite{Waring2020}.

\subsection{Unimodal approaches}

\subsubsection{Tabular}

We implement the tabular component of our multimodal framework using AutoPrognosis 2.0 \cite{Imrie2023AutoPrognosis}.
In contrast to many other approaches for learning from tabular data, we consider full ML \textit{pipelines}, rather than just predictive models, consisting of missing data imputation, feature processing, model selection, and hyperparameter optimization.
AutoPrognosis includes implementations for 22 classification algorithms, nine imputation algorithms, including mean imputation, MICE \cite{mice2011}, and MissForest \cite{stekhoven2011missforest}, five dimensionality reduction, such as PCA, and six feature scaling algorithms.
The best-performing pipelines are then combined into an ensemble via either a learned weighting or stacking, where a meta-model is trained on the output of the underlying pipelines.
For further detail, we refer the reader to Imrie et al. \cite{Imrie2023AutoPrognosis}

\subsubsection{Imaging}

\begin{figure*}[th!]
  \centering
  \includegraphics[width=0.99\textwidth]{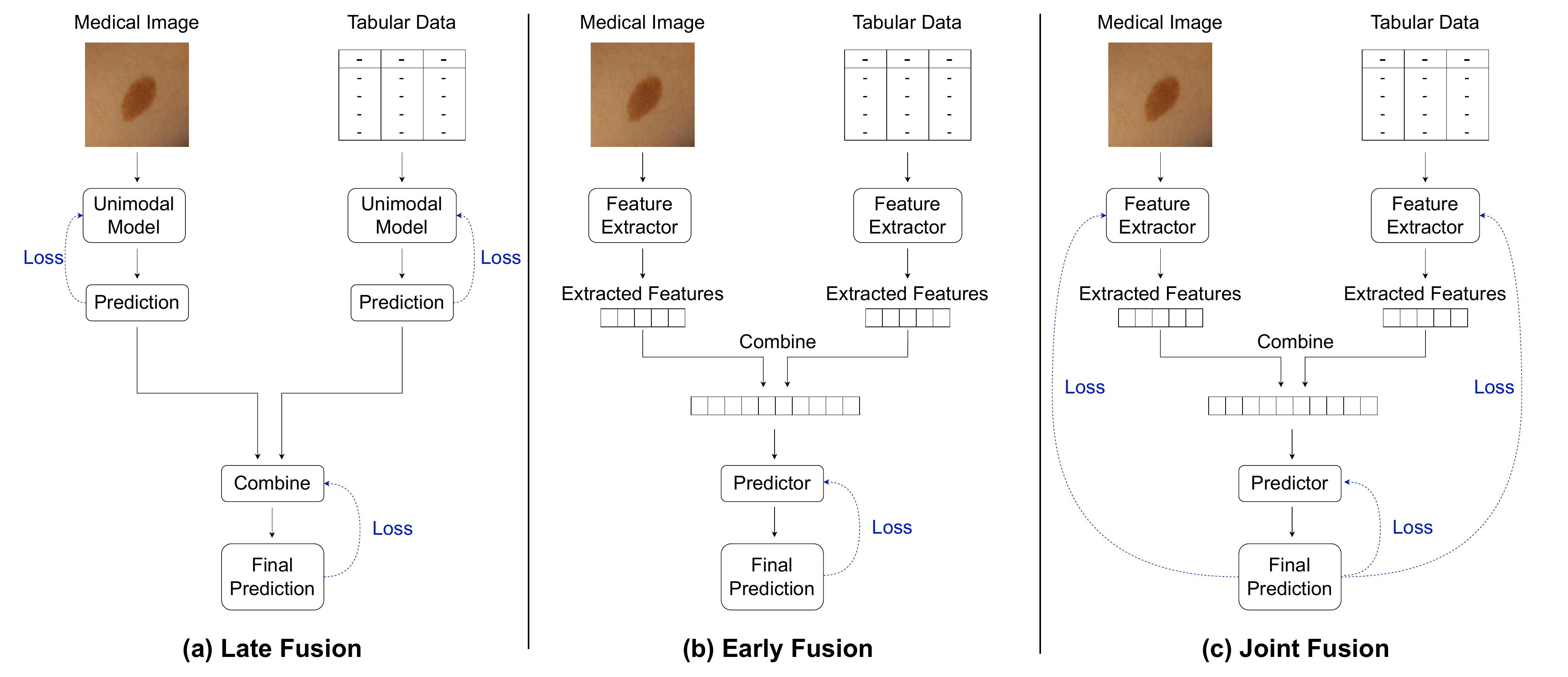}
  \caption{\textbf{Illustration of the three types of multimodal fusion.} (a) Late fusion combines the predictions of separate unimodal models. (b) Early fusion trains a predictive model on the combination of fixed extracted features. (c) Joint fusion flexibly integrates multiple modalities, learning to extract representations and make predictions simultaneously in an end-to-end manner.}
  \label{fig:fusion_types}
\end{figure*} 

For imaging tasks, we employ a several distinct model architectures to cater to a wide range of diagnostic and prognostic applications. 
Specifically, we utilized Convolutional Neural Networks (CNNs), including ResNet \cite{ResNet}, EfficientNet \cite{Efficientnet}, and MobileNetV2 \cite{MobileNetV2}, as well as Vision Transformers (ViTs) \cite{ViT}.
Each model architecture is available in several sizes to be able to handle different task complexities.
A full list of the 17 imaging models provided, together with additional details, can be found in Table \ref{tbl:imaging_models}.

While general-purpose models for tabular data do not exist, the transferability of imaging models has been shown in a diverse range of disciplines \cite{transferlearning2016,transferlearning2020}, and can be particularly effective when there is limited available data for a particular problem.
Pretrained models are especially useful when the available data for a specific task is scarce or when computational resources are limited. 
They allow one to leverage learned features and patterns from vast datasets, potentially improving performance on related tasks.

While most models are pretrained in a supervised manner, self-supervised pretraining has been shown to improve performance on many classification tasks.
Thus, in addition to supervised ViT \cite{ViT}, we also consider DINOv2 \cite{DinoV2}. 
One approach to using pretrained models for new prediction tasks is to extract a fixed representation and train a new classification head on the available task-specific data.
However, often these representations are not well adapted for the specific data under consideration, especially when transferring from the natural image to the medical image domain. 
In this case, we can train these models further by fine-tuning the entire model on the available data.
Fine-tuning is most important when the target task or domain is related but not identical to the one on which the model was originally trained by adapting the generalized capabilities of the pretrained model to specific needs without the necessity of training a model from scratch.
The optimal training strategy depends on the specific task, availability of data, and computational resources.
{\proposed} can be used to train vision models from scratch, use their existing representations, or finetune on the available data.

\subsection{Multimodal data integration}
 
Multimodal ML seeks to train a model that effectively integrates multiple types of data to enable more accurate predictions than can be obtained using any single modality. 
Modalities can exhibit different relationships, including redundancy, depending on the interactions between the information contained in each source \cite{baltruvsaitis2018multimodal}, which can additionally vary in complexity.
Thus a key challenge is discerning the relation between modalities and learning how to integrate modalities most effectively.

We can typically decompose multimodal architectures into two components: modality-specific representations and a joint prediction \cite{baltruvsaitis2018multimodal}. 
Multimodal learning strategies differ primarily in the nature of these components and whether they are jointly or separately learned.
Three main multimodal strategies exist and are incorporated in {\proposed}: Late Fusion, Early Fusion, and Joint Fusion (Fig. \ref{fig:fusion_types}).

\subsubsection{Late Fusion}
Late fusion is an ensemble-based approach that combines predictions from multiple unimodal models, and thus is sometimes referred to as decision-level fusion \cite{sharma1998toward} (Fig. \ref{fig:fusion_types}a).
Each modality is processed independently using a modality-specific model before the predictions are combined.
This allows the user to find the best classifier for each modality independently and evaluate whether each modality has predictive power for the original task. 
However, this has the drawback of not permitting interactions between modalities before the final output, which could result in suboptimal performance when the relationship between modalities is crucial for making accurate predictions or decisions.
One benefit of late fusion is the ability to incorporate new modalities by adding an additional unimodal model and retraining only the ensembling step.
We implement late fusion using a weighted combination of unimodal tabular and imaging models.

\subsubsection{Early fusion}

Early fusion necessitates the translation of each modality into a representation that can be combined using a fusion module, such as concatenation, into a single, unified representation (Fig. \ref{fig:fusion_types}b). 
The combined representation is then used as input to train a separate predictive model.

Compared to late fusion, early fusion allows interactions between different modalities to be captured by the predictive model.
However, the representations are fixed and translating different data types into effective (latent) representations to form a unified representation can be challenging, especially when dealing with heterogeneous data sources with differing scales, dimensions, or types of information.

For image data, a common strategy is to use an intermediate (or the last) layer of a vision model that was either trained to solve the relevant prediction task or a general-purpose imaging model.
For tabular data, especially if the number of features is relatively modest, the representation step can be skipped and the original features are directly combined with the latent representations extracted from the other modalities. 
We used concatenation to combine the imaging and tabular features and trained a fully connected neural network on this representation.

\subsubsection{Joint Fusion}

The fixed, independent representations used in early fusion may not capture relevant factors for the joint prediction task. 
Joint fusion \cite{Huang2020} (or intermediate fusion \cite{stahlschmidt2022multimodal}) aims to improve these representations using joint optimization to enable both cross-modal relationships and modality-specific features to be learned using end-to-end training (Fig. \ref{fig:fusion_types}c). 
The added flexibility of joint fusion can come at the cost of potentially overfitting, especially in the limited data setting.

The most popular approaches for joint fusion use differentiable unimodal models to produce latent representations that are combined via concatenation and passed to a prediction head. 
The system is then trained in an end-to-end manner, either from scratch or using pre-trained unimodal models.
We implement joint fusion similarly to early fusion, except we train end-to-end.

\subsection{Fusion Ensembles}

One major challenge is determining which fusion approach is best. Further, no individual fusion approach may be universally best for all patients.
Ensembling has repeatedly been shown to lead to improved model performance, even when multiple copies of the same model are trained, but can be particularly beneficial when different approaches are combined due to the increased diversity of predictions \cite{krogh1994neural}.
The three fusion approaches learn to combine the information from multiple modalities in distinct ways.
Thus combining the strengths of these different strategies via an ensembling approach could improve both the absolute performance and the robustness of the final model. 
We combine the best-performing unimodal and multimodal fusion approaches in a weighted ensemble.
All ensembles were determined using Bayesian optimization \cite{akiba2019optuna}.
We refer to this ensemble as {\proposed} in our experiments.

\subsection{Explainability}

Models must be thoroughly understood and debugged to validate the underlying logic of the model, engender model trust from both clinical users and patients \cite{Imrie2023}, and satisfy regulatory requirements prior to clinical use \cite{MHRA2023}. 
This is particularly true in the multimodal setting, where we wish to understand what information is being used from each modality and for which patients each modality is contributing to the model output.
Consequently, {\proposed} contains multiple classes of explainability techniques to enable ML models to be better understood. 
We have included feature-based interpretability methods, such as SHAP \cite{lundberg2017unified} and Integrated Gradients \cite{sundararajan2017axiomatic}, that allow us to understand the importance of individual features, as well as an example-based interpretability method, SimplEx \cite{Crabbe2021Simplex}, that explains the model output for a particular sample with examples of similar instances, similar to case-based reasoning. 

\subsection{Uncertainty estimation}

Quantifying the uncertainty of predictions is another critical component in engendering model trust with both clinicians and patients, and can be used both to protect against likely inaccurate predictions and inform clinical decision-making \cite{Kompa2021,Helou2020}. 
We adopted the conformal prediction framework for uncertainty quantification. 
Conformal prediction produces statistically valid prediction intervals or sets for any underlying predictor while making minimal assumptions \cite{vovk2005algorithmic}.
We used inductive conformal prediction \cite{vovk2012conditional}, which uses a calibration set to determine the width of prediction intervals or the necessary size of the prediction sets, with local adaptivity to adjust the interval or set to the specific example \cite{papadopoulos2011regression,johansson2015efficient,seedat2023improving}. For our experiments, we used regularized adaptive prediction sets \cite{angelopoulos2020uncertainty}.

\section{Experiments}

We demonstrate the application of {\proposed} to multimodal healthcare data with the example of skin lesion diagnosis.
This task is inherently a multimodal process: primary care physicians, dermatologists, or other clinicians use multiple factors to determine a diagnosis. 
Visual inspection has formed a crucial element in lesion diagnosis, for example the ``ABCD'' rule or the ELM 7-point checklist \cite{argenziano1998epiluminescence}. These approaches have been refined to include other characteristics beyond the appearance of the lesion at a single point in time, such as evolution \cite{abbasi2004early}. Beyond visual examination, clinicians also consider medical history and other factors, such as itching and bleeding \cite{Bergenmar2002}.

\subsection{Data}

Experiments were conducted using PAD-UFES-20 \cite{pacheco2020pad}. The dataset contains 2,298 skin lesion images from 1,373 patients in Brazil. Images of lesions were captured from smartphones and each image is associated with 21 tabular features, including the patient's age, the anatomical region where the lesion is located, demographic information, and other characteristics of the legion, such as whether it itched, bled, or had grown.
An overview of the clinical features can be found in Table \ref{tbl:clinical_variables}. 
Further details can be found in the publication describing the dataset \cite{pacheco2020pad}.

\begin{table*}[ht]
\caption{\textbf{Unimodal skin legion classification performance.} The best result for each modality is in bold. The best non-ensemble approach for each modality is underlined.}
\vspace{0.15cm}
\label{tbl:unimodal_results}
\centering
\small
\resizebox{\textwidth}{!}{
\begin{tabular}{lcccc|cccc}
\toprule
                    & \multicolumn{4}{c}{\textbf{Lesion Categorization (6-way)}} & \multicolumn{4}{c}{\textbf{Cancer Diagnosis (Binary)}} \\
\textbf{Method}     & \textbf{Acc.} & \textbf{Bal. Acc.} & \textbf{AUROC} & \textbf{F1} & \textbf{Acc.} & \textbf{AUROC} & \textbf{F1} & \textbf{MCC} \\
\midrule
\textbf{Tabular} \\
Log. Reg.       & 63.6\% & \underline{\textbf{63.3\%}} & \underline{0.890} & \underline{0.559} & 83.0\% & \underline{0.904} & 0.814 & 0.657 \\
Random Forest   & 65.2\% & 54.0\% & 0.865 & 0.535 & 83.0\% & 0.903 & 0.810 & 0.662 \\
XGBoost         & 66.5\% & 54.4\% & 0.875 & 0.545 & 81.3\% & 0.885 & 0.797 & 0.623 \\
CatBoost        & 64.3\% & 57.2\% & 0.877 & 0.545 & \underline{83.4\%} & 0.902 & \underline{0.822} & \underline{0.667} \\
MLP             & \underline{\textbf{69.7\%}} & 52.8\% & 0.878 & 0.526 & 83.1\% & 0.902 & 0.819 & 0.663 \\
AutoPrognosis   & 69.4\% & 61.9\% & \textbf{0.891} & \textbf{0.580} & \textbf{83.9\%} & \textbf{0.909} & \textbf{0.825} & \textbf{0.676} \\
\midrule
\textbf{Imaging} \\
ResNet18 & 59.8\% & 57.8\% & 0.885 & 0.547 & 81.9\% & 0.897 & 0.808 & 0.637 \\
ResNet34 & 57.4\% & 54.5\% & 0.873 & 0.517 & 80.3\% & 0.881 & 0.790 & 0.603 \\
ResNet50 & 60.7\% & 60.0\% & 0.888 & 0.562 & 82.0\% & 0.888 & 0.811 & 0.637 \\
ResNet101 & 60.3\% & 56.6\% & 0.886 & 0.543 & 81.6\% & 0.883 & 0.802 & 0.628 \\
ResNet152 & 63.6\% & 59.6\% & 0.895 & 0.578 & 82.1\% & 0.892 & 0.810 & 0.640 \\
EfficientNetB0 & 64.3\% & 60.8\% & 0.899 & 0.577 & 82.4\% & 0.900 & 0.807 & 0.645 \\
EfficientNetB1 & 65.5\% & 63.7\% & 0.901 & 0.602 & 82.6\% & 0.899 & 0.811 & 0.648 \\
EfficientNetB2 & 65.1\% & 59.7\% & 0.899 & 0.578 & 81.5\% & 0.888 & 0.801 & 0.628 \\
EfficientNetB3 & 64.2\% & 62.6\% & 0.902 & 0.598 & 81.9\% & 0.898 & 0.805 & 0.635 \\
EfficientNetB4 & 66.7\% & 62.1\% & 0.899 & 0.602 & 81.9\% & 0.897 & 0.807 & 0.635 \\
EfficientNetB5 & 66.7\% & 62.8\% & 0.904 & 0.609 & 82.6\% & 0.903 & 0.810 & 0.649 \\
MobileNetV2 & 58.4\% & 54.5\% & 0.868 & 0.512 & 79.6\% & 0.877 & 0.779 & 0.590 \\
ViTBase & 67.1\% & 65.0\% & \underline{0.917} & 0.618 & 82.9\% & 0.913 & 0.816 & 0.657 \\
ViTLarge & 68.1\% & 65.2\% & 0.916 & 0.631 & 84.1\% & \underline{0.916} & 0.831 & 0.682 \\
DinoV2Small & 68.1\% & 65.0\% & 0.913 & 0.630 & 84.2\% & 0.912 & \underline{0.834} & 0.683 \\
DinoV2Base & 68.5\% & \underline{65.9\%} & 0.914 & 0.639 & 84.0\% & 0.913 & 0.833 & 0.679 \\
DinoV2Large & \underline{69.0\%} & 65.8\% & 0.916 & \underline{0.640} & \underline{84.4\%} & 0.913 & \underline{0.834} & \underline{0.686} \\
Imaging ensemble & \textbf{71.6\%} & \textbf{69.4\%} & \textbf{0.927} & \textbf{0.672} & \textbf{85.3\%} & \textbf{0.927} & \textbf{0.845} & \textbf{0.706} \\
\bottomrule
\end{tabular}
}
\vspace{0.15cm}
\end{table*}

Skin lesions are classified as one of six different diagnoses, three of which are cancerous (Basal Cell Carcinoma, Squamous Cell Carcinoma, and Melanoma) and three are non-cancerous (Actinic Keratosis, Melanocytic Nevus, and Seborrheic Keratosis).
As is common in studies of skin lesions, there are substantial differences in the number of lesions with each diagnosis, resulting in class imbalance (Table \ref{tbl:clinical_variables}).
Aggregating the diagnoses into cancerous and non-cancerous almost eliminates this imbalance (47\% cancerous, 53\% non-cancerous).
To demonstrate the suitability of our framework for balanced and imbalanced classification scenarios, we explore both the task of predicting the specific diagnoses and the binary determination of whether a given lesion is cancerous.
We assessed lesion categorization usin accuracy, balanced accuracy, area under the receiver operating curve (AUROC), and macro F1 score, and assessed cancer diagnosis using accuracy, AUROC, F1 score and Matthew's correlation coefficient (MCC).

To account for the presence of multiple images for some patients and the imbalance in incidence of different diagnoses, we conducted 5-fold cross-validation with stratified sampling, ensuring all images from the same patient were contained in the same fold. Additionally, we used 20\% of the training set of each fold to optimize hyperparameters and ensemble weights.
13 clinical variables had substantial levels of missingness (c. 35\%) and this missingness was often strongly associated with diagnosis. 
Consequently, we retained only features without this missingness. Some other features had entries recorded as ``unknown,'' corresponding to the patient being asked the question but not knowing the answer. This was the case for six features, with an occurrence between 0.1\% and 17\%. 
This resulted in eight tabular variables being retained.
Categorical variables were one-hot encoded, yielding 27 clinical features and one image for each sample.
 
\subsection{How predictive is each modality in isolation?}

Collecting additional information is never without expense. This can be financial, time, or even adverse effects of collecting the information.
Thus it is critical to understand whether a modality is necessary and brings additional predictive power.
Therefore, we first used {\proposed} to optimize ML pipelines for each modality separately to quantify the value of imaging and clinical features individually.
Both the clinical variables and lesion images exhibit some predictive power for lesion categorization and cancer diagnosis (Table \ref{tbl:unimodal_results}), with the individual imaging models outperforming the tabular models for lesion categorization while achieving similar results for cancer diagnosis.

\textbf{Tabular.} We tested several classification models in isolation, as well as the performance of {\proposed} on just the tabular modality, which is equivalent to AutoPrognosis \cite{Imrie2023AutoPrognosis}.
Overall, AutoPrognosis outperformed any individual tabular model, in particular for cancer diagnosis, demonstrating the importance of AutoML and ensembling (Table \ref{tbl:unimodal_results}). However, the relative outperformance over logistic regression for lesion categorization and CatBoost for cancer diagnosis was relatively minor, perhaps reflecting the nature of the structured information available.

\textbf{Imaging.} The different imaging architectures displayed significant variability in performance across the lesion categorization and cancer diagnosis prediction tasks, however several trends could be observed. 
The vision transformer architectures (ViT and DINOv2) outperformed the CNN-based models (ResNet, EfficientNet, MobileNet) almost universally across both tasks. One explanation beyond the model architecture could be the pre-training set, which differed between the transformer and CNN models (see Table \ref{tbl:imaging_models}).
Increasing the size of models typically led to improvements in performance for the transformer models, although the largest models did not necessarily improve performance (e.g. DINOv2Large vs. DINOv2Base), while the trend was less clear for the CNN-based models.
All model architectures consistently underperformed when trained from initialization, thus all results shown are from fine-tuning pre-trained models.

Ensembling the best-performing image models resulted in a substantial increase in performance across all metrics for both prediction tasks. While the transformers outperformed the CNNs individually, many of the ensembles contained CNN-based approaches, demonstrating the importance of diversity in ensemble learning.

\begin{table*}[ht]
\caption{\textbf{Skin legion classification performance.} All multimodal approaches outperform the unimodal baselines. {\proposed} achieves the best results.}
\vspace{0.15cm}
\label{tbl:main_results}
\centering
\small
\resizebox{\textwidth}{!}{
\begin{tabular}{lcccc|cccc}
\toprule
 & \multicolumn{4}{c}{\textbf{Lesion Categorization (6-way)}} & \multicolumn{4}{c}{\textbf{Cancer Diagnosis (Binary)}} \\
\textbf{Method}      & \textbf{Acc.} & \textbf{Bal. Acc.} & \textbf{AUROC} & \textbf{F1} & \textbf{Acc.} & \textbf{AUROC} & \textbf{F1} & \textbf{MCC} \\
\midrule
Tabular ensemble      & 69.4\% & 61.9\% & 0.891 & 0.580 & 83.9\% & 0.909 & 0.825 & 0.676 \\ 
Best image model      & 68.5\% & 65.9\% & 0.914 & 0.639 & 84.4\% & 0.913 & 0.834 & 0.686 \\ 
Imaging ensemble      & 71.6\% & 69.4\% & 0.927 & 0.672 & 85.3\% & 0.927 & 0.845 & 0.706 \\ 
\midrule
Best late fusion      & 77.0\% & 72.8\% & 0.935 & 0.706 & 89.2\% & 0.950 & 0.883 & 0.782 \\ 
Late fusion ensemble  & 79.0\% & 74.7\% & 0.939 & 0.730 & 89.1\% & 0.954 & 0.882 & 0.781 \\ 
Best early fusion     & 70.9\% & 68.1\% & 0.918 & 0.657 & 85.7\% & 0.922 & 0.847 & 0.713 \\ 
Early fusion ensemble & 74.7\% & 70.2\% & 0.930 & 0.701 & 86.7\% & 0.936 & 0.856 & 0.731 \\ 
Best joint fusion     & 73.8\% & 71.4\% & 0.930 & 0.698 & 87.5\% & 0.940 & 0.866 & 0.748 \\ 
Joint fusion ensemble & 75.6\% & 71.9\% & 0.937 & 0.716 & 88.8\% & 0.951 & 0.880 & 0.775 \\
\midrule
\textbf{{\proposed}}  & \textbf{80.6\%} & \textbf{75.8\%} & \textbf{0.945} & \textbf{0.758} & \textbf{89.8\%} & \textbf{0.956} & \textbf{0.889} & \textbf{0.794} \\
\bottomrule
\end{tabular}
}
\vspace{0.15cm}
\end{table*}

\subsection{What benefit does multimodal ML provide?}

Now that we have shown that each modality has a potential predictive capability for skin lesion diagnoses, we sought to quantify what, if any, benefit incorporating both modalities when issuing predictions provides. To do this, we assessed each of the multimodal strategies included in {\proposed}. 

All three fusion approaches demonstrate significant improvements over the unimodal classifiers, demonstrating the importance of integrating data from multiple sources. In Table \ref{tbl:main_results}, we report the best single model for each fusion strategy, together with the impact of ensembling the best-performing models as measured on the held-out portion of the training set. The impact of combining the modalities varied across the various model architectures, with the results also differing for each of the fusion strategies, late (Table \ref{tbl:late_fusion_results}), early (Table \ref{tbl:early_fusion_results}), and joint (Table \ref{tbl:joint_fusion_results}).

Perhaps surprisingly, late fusion outperformed both early and joint fusion, with early fusion the worst-performing fusion approach. This is likely a consequence of the relatively strong predictive power of the tabular features and the number of samples available, but could also reveal the nature of the relationship between the two modalities.
Again, ensembling the best-performing models for each fusion strategy provides a relatively small but consistent improvement, except for the cancer diagnosis task for late fusion, where the best individual model performed similarly to the ensemble.

{\proposed} leverages the power of each fusion strategy and the unimodal models by combining them in an ensemble. This approach performed best across both tasks as measured by any metric, improving the performance over any one fusion approach alone. Despite late fusion outperforming the other multimodal and unimodal approaches, it was not always selected as the most important component in the ensemble and the largest weight assigned to any of the five strategies (two unimodal, three fusion) was 39\%, further reinforcing the importance of diversity. 

\begin{figure*}[ht!]
    \centering
    \begin{subfigure}[t]{0.5\textwidth}
        \centering
        \includegraphics[width=\linewidth]{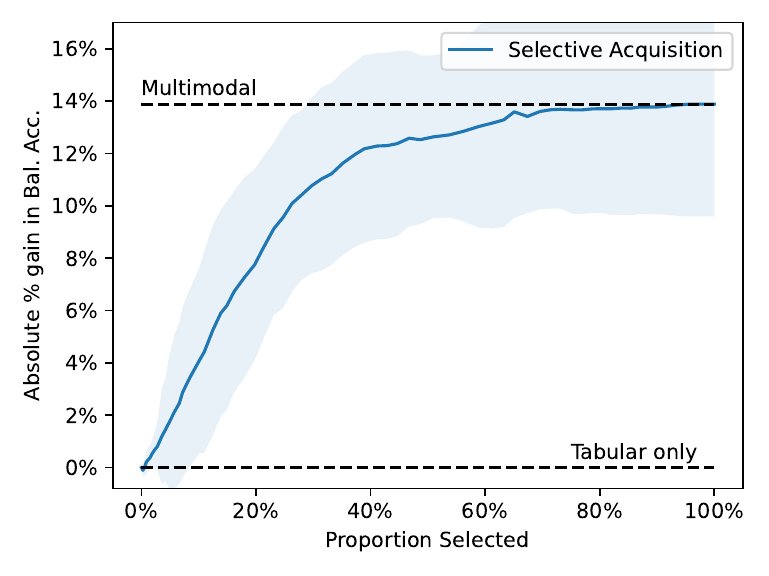}
        \caption{Lesion categorization} \label{fig:uncertainty_lesion}
    \end{subfigure}%
    ~ 
    \begin{subfigure}[t]{0.5\textwidth}
        \centering
        \includegraphics[width=\linewidth]{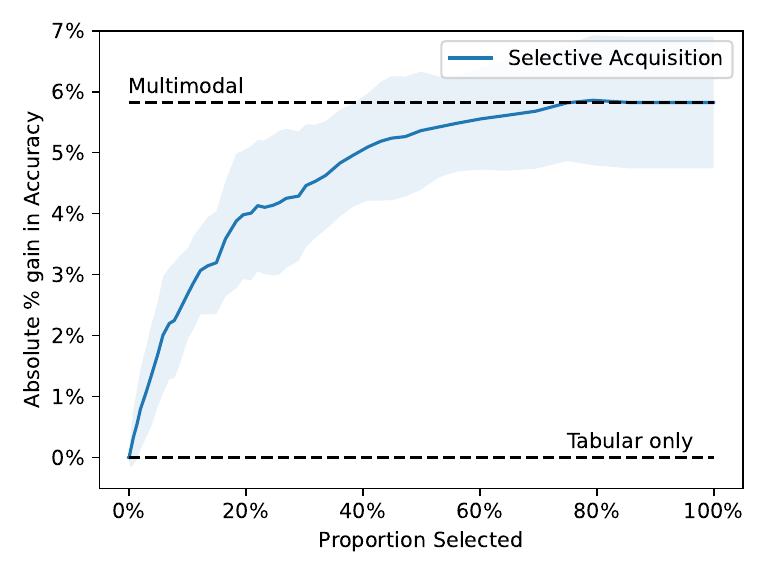}
        \caption{Cancer diagnosis} \label{fig:uncertainty_cancer}
    \end{subfigure}
  \caption{Selective acquisition of images based on conformal prediction. By acquiring images for around 20\% of samples with the highest predicted uncertainty based on the tabular features, we capture c. 55\% and 65\% of the improvement of the multimodal classifier for (a) lesion categorization and (b) cancer diagnosis, respectively. We approach the performance of the multimodal classifier by acquiring images for around half of all patients.}
  \label{fig:uncertainty}
  \vspace{-0.1cm}
\end{figure*} 

\subsection{When are additional modalities most helpful?}

While we have shown multimodal systems significantly outperform unimodal approaches for lesion categorization and cancer diagnosis, we might not require all modalities for all patients. As mentioned previously, there might be downsides to collecting additional information, thus identifying when we would benefit is both important and often a key clinical decision, since modalities are typically acquired sequentially. 

We demonstrate how {\proposed} can be used to answer this question. We assumed access initially to the clinical features and wanted to identify for which patients to acquire an image of the lesion.
We used conformal prediction to estimate the uncertainty of each prediction and chose to acquire images for the patients with the highest uncertainty.

By acquiring images for around 20\% of samples with the highest predicted uncertainty based on the tabular features, we can capture around two-thirds of the total improvement of the multimodal ensemble classifier for the cancer diagnosis task (Fig. \ref{fig:uncertainty_cancer}) and over half for the lesion categorization task (Fig. \ref{fig:uncertainty_lesion}). Acquiring lesion images for around 50\% of patients results in coming close to matching the performance of the multimodal classifier, thereby halving the number of images needed to be collected.

\begin{figure*}[th!]
  \centering
  \includegraphics[width=\textwidth, trim={6em 2em 3em 6em}, clip]{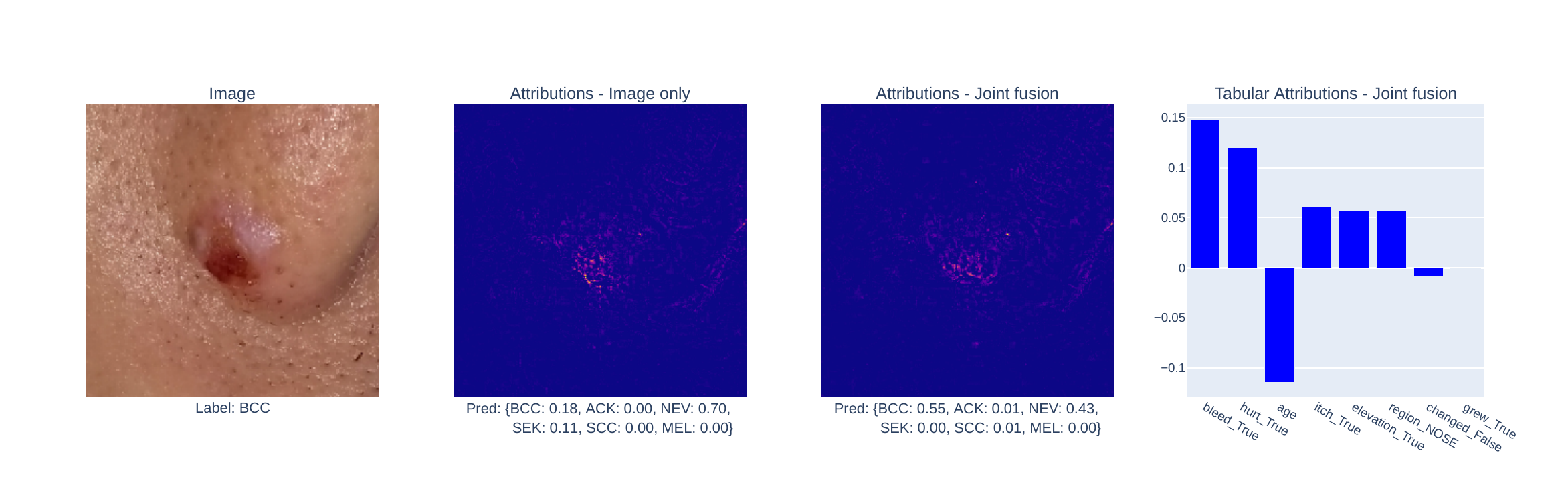}
  \caption{Comparison of explanations for unimodal and multimodal models using integrated gradients. The original image (left, img\_id: PAT\_521\_984\_412) together with attributions for the unimodal (center left) and joint fusion EfficientNetB4 models (center right and right).}
  \label{fig:explainability}
  \vspace{-0.1cm}
\end{figure*} 

\subsection{Understanding the information provided by each modality}

Understanding why predictions were issued is incredibly important across medical contexts. We demonstrate how the interpretability techniques included in {\proposed} can be used to analyze the rationale for predictions across multiple modalities.
We used integrated gradients \cite{sundararajan2017axiomatic} to analyze the predictions from the image-only and joint fusion variants of EfficientNetB4. An example is shown in Fig. \ref{fig:explainability}. 

The image-only model incorrectly classifies the lesion as Melanocytic Nevus (NEV, non-cancerous), while the joint fusion model correctly identifies the lesion as Basal Cell Carcinoma (BCC, cancerous). The image attributions (Fig. \ref{fig:explainability} center) are somewhat similar, both placing the most importance on the lesion, although there are minor differences in several areas. 
Importantly, the clinical variables allow the multimodal approach to correct the image-only prediction. NEV is typically asymptomatic \cite{macneal2020congenital} and more common in younger individuals \cite{Zalaudek2011}. The patient reported the lesion had bled, hurt, and itched, which the multimodal model correctly identified made NEV less likely and increased the chance of BCC, offset by the relatively young age of the patient (32) which reduced the magnitude of the BCC prediction.
This example clearly demonstrates the importance of incorporating both modalities and the understanding that explainability methods can provide.

\section{Discussion}

Predictive modeling has the potential to support clinical decision-making and improve outcomes. However, incorporating multiple types of data into computational approaches is not yet widespread in medicine and healthcare.
In this paper, we demonstrated the utility of {\proposed} for developing clinical models from multimodal data using AutoML.
Our framework simplifies the application of multimodal fusion strategies, automatically determining the best strategy for the available data and clinical application.

We have shown how {\proposed} can be used to perform unimodal analysis for tabular and imaging data, enabling clinicians to understand when multimodal approaches will provide benefit. Beyond prediction, we used uncertainty estimation to determine for which patients additional information is necessary and explainability techniques to improve model understanding.

The use of AutoML frameworks such as {\proposed} can aid in model development, but does not alter the necessity to ensure models are suitably validated to ensure they exhibit the desired characteristics, such as being accurate, reliable, and fair.
As with any learning algorithm, significant care must be taken by the user to ensure appropriate study design and data curation, without which an inaccurate or biased model could be developed which could have adverse effects on patient health. 

While there have been bespoke multimodal ML systems developed, few general-purpose frameworks exist.
HAIM \cite{Soenksen2022} is an early fusion approach using user-defined pre-trained feature-extraction models to extract representations that are concatenated and passed to an XGBoost model. 
Wang et al. proposed a multimodal approach for esophageal variceal bleeding prediction \cite{Wang2023}. They first trained an imaging model and then used automated machine learning to develop a classifier based on structured clinical data and the output of the imaging model.
Finally, AutoGluon-Multimodal \cite{tang2024autogluon} enables fine-tuning of pretrained models across multiple modalities, combining their outputs via late fusion.
In contrast, {\proposed} incorporates multiple fusion approaches, including both early and late fusion as possible strategies, while our experiments highlight the limitations of only considering a single fusion strategy.

While in this paper, we demonstrated the application of {\proposed} to the problem of diagnosing skin lesions using smartphone images and clinical variables, our framework is generally applicable and can naturally be extended to additional modalities and both new unimodal models and fusion strategies.
We believe {\proposed} represents a powerful tool for clinicians and ML experts when working with data from multiple modalities and hope our aids the adoption of multimodal ML methods in healthcare and medicine.

\backmatter

\bmhead{Code and data availability}\label{sec:code_data}
{\proposed} is available at \url{https://github.com/vanderschaarlab/AutoPrognosis-Multimodal}.
PAD-UFES-20 is freely available at \url{https://data.mendeley.com/datasets/zr7vgbcyr2/1} (DOI: 10.17632/zr7vgbcyr2.1) \cite{pacheco2020pad}.
All preprocessing and the splits used for our experiments can be found at \url{https://github.com/vanderschaarlab/AutoPrognosis-Multimodal}.

\bibliography{bibliography}

\clearpage

\begin{appendices}

\renewcommand{\thefigure}{S.\arabic{figure}}
\setcounter{figure}{0}

\renewcommand{\thetable}{S.\arabic{table}}
\setcounter{table}{0}

\begin{table}[ht!]
\caption{\textbf{Imaging models included in {\proposed}.} CNN - Convolutional Neural Network. ViT - Vision Transformer.}
\label{tbl:imaging_models}
\centering
\begin{tabular}{cccccc}
\toprule
\textbf{Model} & \textbf{Type} & \textbf{\# Param.} & \makecell{\textbf{Pretraining} \\ \textbf{Data}} & \makecell{\textbf{Embedding} \\ \textbf{size}} & \textbf{Ref.} \\ 
\midrule
ResNet18 & CNN & 11.7 M & ImageNet-1k & 512 & \cite{ResNet} \\
ResNet34 & CNN & 21.8 M & ImageNet-1k & 512 & \cite{ResNet} \\
ResNet50 & CNN & 25 M & ImageNet-1k & 2048 & \cite{ResNet} \\
ResNet101 & CNN & 44.5 M & ImageNet-1k & 2048 & \cite{ResNet} \\
ResNet152 & CNN & 60.2 M & ImageNet-1k & 2048 & \cite{ResNet} \\
EfficientNetB0 & CNN & 5.3 M & ImageNet-1k & 320 & \cite{Efficientnet} \\
EfficientNetB1 & CNN & 7.8 M & ImageNet-1k & 320 & \cite{Efficientnet} \\
EfficientNetB2 & CNN & 9.2 M & ImageNet-1k & 352 & \cite{Efficientnet} \\
EfficientNetB3 & CNN & 12 M & ImageNet-1k & 384 & \cite{Efficientnet} \\
EfficientNetB4 & CNN & 19 M & ImageNet-1k & 448 & \cite{Efficientnet} \\
EfficientNetB5 & CNN & 30 M & ImageNet-1k & 512 & \cite{Efficientnet} \\
MobileNetV2 & CNN & 3.4 M & ImageNet-1k & 320 & \cite{MobileNetV2} \\
ViTBase & ViT-B/16 & 86 M & ImageNet-1k & 768 & \cite{ViT} \\
ViTLarge & ViT-L/16 & 307 M & ImageNet-21k & 1024 & \cite{ViT} \\
DinoV2Small & ViT-S/14 & 22 M & LVD-142M & 384 & \cite{DinoV2} \\
DinoV2Base & ViT-B/14 & 86 M & LVD-142M & 768 & \cite{DinoV2} \\
DinoV2Large & ViT-L/14 & 307 M & LVD-142M & 1024 & \cite{DinoV2} \\ 
\bottomrule
\end{tabular}
\vspace{-0.1in}

\end{table}

\begin{table}[ht!]
\caption{Clinical variables in the PAD-UFES-20 dataset (n=2,298).}
\label{tbl:clinical_variables}
\centering
\begin{tabular}{lc}
\toprule
\textbf{Diagnosis}            &                \\
Basal Cell Carcinoma (BCC)    & 845 (36.8\%)   \\
Squamous Cell Carcinoma (SCC) & 192 (8.4\%)    \\
Melanoma (MEL)                & 52 (2.3\%)     \\
Actinic Keratosis (ACK)       & 730 (31.8\%)   \\
Melanocytic Nevus (NEV)       & 244 (10.6\%)   \\
Seborrheic Keratosis (SEK)    & 235 (10.2\%)   \\
\textbf{Age}            &                \\
6-29                    & 92 (4.0\%)     \\
30-49                   & 386 (16.8\%)   \\
50-69                   & 1,098 (47.8\%) \\
70-94                   & 722 (31.4\%)   \\
\textbf{Region}         &                \\
Face                    & 570 (24.5\%)   \\
Forearm                 & 392 (17.1\%)   \\
Chest                   & 280 (12.2\%)   \\
Back                    & 248 (10.8\%)   \\
Arm                     & 192 (8.4\%)    \\
Nose                    & 158 (6.9\%)    \\
Hand                    & 126 (5.5\%)    \\
Neck                    & 93 (4.0\%)     \\
Thigh                   & 73 (3.2\%)     \\
Ear                     & 73 (3.2\%)     \\
Abdomen                 & 36 (1.6\%)     \\
Lip                     & 23 (1.0\%)     \\
Scalp                   & 18 (0.8\%)     \\
Foot                    & 16 (0.7\%)     \\
\textbf{Itch}           &                \\
Yes                     & 1,455 (63.3\%) \\
No                      & 837 (36.4\%)   \\
Unknown                 & 6 (0.3\%)      \\
\textbf{Grew}           &                \\
Yes                     & 925 (40.2\%)   \\
No                      & 971 (42.3\%)   \\
Unknown                 & 402 (17.5\%)   \\
\textbf{Hurt}           &                \\
Yes                     & 397 (17.3\%)   \\
No                      & 1,891 (82.3\%) \\
Unknown                 & 10 (0.4\%)     \\
\textbf{Changed}        &                \\
Yes                     & 202 (0.9\%)    \\
No                      & 1,700 (74.0\%) \\
Unknown                 & 396 (17.2\%)   \\
\textbf{Bleed}          &                \\
Yes                     & 614 (26.7\%)   \\
No                      & 1,678 (73.0\%) \\
Unknown                 & 6 (0.3\%)      \\
\textbf{Elevation}      &                \\
Yes                     & 1,433 (62.4\%) \\
No                      & 863 (37.6\%)   \\
Unknown                 & 2 (0.1\%)      \\
\bottomrule
\end{tabular}
\vspace{-0.1in}
\end{table}

\begin{table*}[ht]
\caption{\textbf{Late fusion skin legion classification performance.} The best result for each modality is in bold. The best non-ensemble approach for each modality is underlined.}
\vspace{0.15cm}
\label{tbl:late_fusion_results}
\centering
\small
\resizebox{\textwidth}{!}{
\begin{tabular}{lcccc|cccc}
\toprule
                    & \multicolumn{4}{c}{\textbf{Lesion Categorization (6-way)}} & \multicolumn{4}{c}{\textbf{Cancer Diagnosis (Binary)}} \\
\textbf{Method}     & \textbf{Acc.} & \textbf{Bal. Acc.} & \textbf{AUROC} & \textbf{F1} & \textbf{Acc.} & \textbf{AUROC} & \textbf{F1} & \textbf{MCC} \\
\midrule
ResNet18 & 74.0\% & 68.0\% & 0.921 & 0.656 & 87.8\% & 0.946 & 0.867 & 0.755 \\
ResNet34 & 73.2\% & 67.1\% & 0.920 & 0.638 & 87.7\% & 0.946 & 0.867 & 0.753 \\
ResNet50 & 73.6\% & 69.8\% & 0.925 & 0.674 & 87.8\% & 0.943 & 0.867 & 0.754 \\
ResNet101 & 73.6\% & 68.4\% & 0.926 & 0.665 & 86.8\% & 0.942 & 0.855 & 0.734 \\
ResNet152 & 75.5\% & 70.2\% & 0.928 & 0.684 & 88.1\% & 0.944 & 0.870 & 0.760 \\
EfficientNetB0 & 75.2\% & 69.3\% & 0.925 & 0.665 & 87.9\% & 0.947 & 0.867 & 0.756 \\
EfficientNetB1 & 75.5\% & 71.3\% & 0.929 & 0.693 & 88.1\% & 0.949 & 0.870 & 0.762 \\
EfficientNetB2 & 75.5\% & 69.4\% & 0.927 & 0.668 & 87.6\% & 0.944 & 0.864 & 0.751 \\
EfficientNetB3 & 74.7\% & 71.0\% & 0.928 & 0.685 & 88.1\% & 0.946 & 0.869 & 0.760 \\
EfficientNetB4 & 75.4\% & 69.9\% & 0.928 & 0.672 & 88.0\% & 0.946 & 0.868 & 0.758 \\
EfficientNetB5 & 76.9\% & 72.0\% & 0.928 & 0.696 & 87.9\% & 0.948 & 0.867 & 0.757 \\
MobileNetV2 & 72.2\% & 65.6\% & 0.912 & 0.627 & 87.1\% & 0.939 & 0.857 & 0.739 \\
ViTBase & 76.6\% & 71.6\% & \underline{0.937} & 0.700 & 87.3\% & 0.948 & 0.861 & 0.744 \\
ViTLarge & 76.3\% & 71.2\% & 0.936 & 0.697 & 88.0\% & 0.950 & 0.869 & 0.759 \\
DinoV2Small & 76.8\% & 70.9\% & \underline{0.937} & 0.695 & \underline{\textbf{89.2\%}} & 0.950 & \underline{\textbf{0.883}} & \underline{\textbf{0.782}} \\
DinoV2Base & 77.0\% & \underline{72.8\%} & 0.935 & \underline{0.706} & 88.5\% & \underline{0.951} & 0.875 & 0.767 \\
DinoV2Large & \underline{77.7\%} & 72.3\% & 0.935 & 0.697 & 87.5\% & 0.948 & 0.864 & 0.749 \\
\midrule
Ensemble & \textbf{79.0\%} & \textbf{74.7\%} & \textbf{0.939} & \textbf{0.730} & 89.1\% & \textbf{0.954} & 0.882 & 0.781 \\
\bottomrule
\end{tabular}
}
\vspace{0.15cm}
\end{table*}

\begin{table*}[ht]
\caption{\textbf{Early fusion skin legion classification performance.} The best result for each modality is in bold. The best non-ensemble approach for each modality is underlined.}
\vspace{0.15cm}
\label{tbl:early_fusion_results}
\centering
\small
\resizebox{\textwidth}{!}{
\begin{tabular}{lcccc|cccc}
\toprule
                    & \multicolumn{4}{c}{\textbf{Lesion Categorization (6-way)}} & \multicolumn{4}{c}{\textbf{Cancer Diagnosis (Binary)}} \\
\textbf{Method}     & \textbf{Acc.} & \textbf{Bal. Acc.} & \textbf{AUROC} & \textbf{F1} & \textbf{Acc.} & \textbf{AUROC} & \textbf{F1} & \textbf{MCC} \\
\midrule
ResNet18 & 59.9\% & 59.2\% & 0.879 & 0.558 & 81.8\% & 0.899 & 0.803 & 0.635 \\
ResNet34 & 58.1\% & 56.3\% & 0.865 & 0.541 & 80.9\% & 0.884 & 0.795 & 0.614 \\
ResNet50 & 62.8\% & 61.7\% & 0.891 & 0.590 & 84.1\% & 0.908 & 0.832 & 0.679 \\
ResNet101 & 66.8\% & 63.4\% & 0.904 & 0.617 & 84.4\% & 0.921 & 0.833 & 0.685 \\
ResNet152 & 69.1\% & 64.6\% & 0.910 & 0.636 & 83.3\% & 0.904 & 0.821 & 0.663 \\
EfficientNetB0 & 66.0\% & 61.2\% & 0.884 & 0.601 & 83.4\% & 0.903 & 0.822 & 0.665 \\
EfficientNetB1 & 64.3\% & 60.8\% & 0.884 & 0.586 & 82.8\% & 0.903 & 0.816 & 0.655 \\
EfficientNetB2 & 63.8\% & 58.7\% & 0.884 & 0.577 & 80.3\% & 0.882 & 0.786 & 0.602 \\
EfficientNetB3 & 64.1\% & 58.8\% & 0.882 & 0.569 & 81.6\% & 0.896 & 0.801 & 0.629 \\
EfficientNetB4 & 66.4\% & 62.2\% & 0.894 & 0.607 & 82.9\% & 0.902 & 0.818 & 0.656 \\
EfficientNetB5 & 66.3\% & 61.2\% & 0.893 & 0.603 & 82.0\% & 0.898 & 0.807 & 0.638 \\
MobileNetV2 & 64.4\% & 59.4\% & 0.880 & 0.576 & 82.1\% & 0.898 & 0.808 & 0.641 \\
ViTBase & 58.1\% & 55.2\% & 0.820 & 0.517 & 80.7\% & 0.881 & 0.792 & 0.612 \\
ViTLarge & 70.1\% & 66.6\% & 0.915 & 0.653 & 84.1\% & 0.914 & 0.829 & 0.680 \\
DinoV2Small & \underline{70.9\%} & \underline{68.1\%} & \underline{0.918} & \underline{0.657} & \underline{85.7\%} & \underline{0.922} & \underline{0.847} & \underline{0.713} \\
DinoV2Base & 69.1\% & 64.8\% & 0.904 & 0.635 & 83.5\% & 0.912 & 0.821 & 0.669 \\
DinoV2Large & 70.4\% & 62.8\% & 0.906 & 0.630 & 83.8\% & 0.915 & 0.829 & 0.674 \\
\midrule
Ensemble & \textbf{74.7\%} & \textbf{70.2\%} & \textbf{0.930} & \textbf{0.701} & \textbf{86.7\%} & \textbf{0.936} & \textbf{0.856} & \textbf{0.731} \\
\bottomrule
\end{tabular}
}
\vspace{0.15cm}
\end{table*}

\begin{table*}[ht]
\caption{\textbf{Joint fusion skin legion classification performance.} The best result for each modality is in bold. The best non-ensemble approach for each modality is underlined.}
\vspace{0.15cm}
\label{tbl:joint_fusion_results}
\centering
\small
\resizebox{\textwidth}{!}{
\begin{tabular}{lcccc|cccc}
\toprule
                    & \multicolumn{4}{c}{\textbf{Lesion Categorization (6-way)}} & \multicolumn{4}{c}{\textbf{Cancer Diagnosis (Binary)}} \\
\textbf{Method}     & \textbf{Acc.} & \textbf{Bal. Acc.} & \textbf{AUROC} & \textbf{F1} & \textbf{Acc.} & \textbf{AUROC} & \textbf{F1} & \textbf{MCC} \\
\midrule
ResNet18 & 65.8\% & 61.6\% & 0.899 & 0.588 & 83.4\% & 0.911 & 0.823 & 0.667 \\
ResNet34 & 62.8\% & 61.6\% & 0.889 & 0.575 & 82.1\% & 0.891 & 0.809 & 0.639 \\
ResNet50 & 66.1\% & 60.1\% & 0.879 & 0.568 & 84.3\% & 0.909 & 0.833 & 0.684 \\
ResNet101 & 67.5\% & 61.5\% & 0.892 & 0.595 & 83.9\% & 0.911 & 0.827 & 0.675 \\
ResNet152 & 68.8\% & 66.3\% & 0.913 & 0.642 & 85.3\% & 0.921 & 0.842 & 0.704 \\
EfficientNetB0 & 62.9\% & 64.0\% & 0.899 & 0.605 & 84.7\% & 0.921 & 0.833 & 0.691 \\
EfficientNetB1 & 66.1\% & 65.1\% & 0.905 & 0.624 & 85.4\% & 0.926 & 0.843 & 0.705 \\
EfficientNetB2 & 68.3\% & 63.9\% & 0.912 & 0.620 & 84.0\% & 0.914 & 0.828 & 0.677 \\
EfficientNetB3 & 68.1\% & 65.3\% & 0.913 & 0.634 & 84.8\% & 0.920 & 0.837 & 0.693 \\
EfficientNetB4 & 69.0\% & 66.2\% & 0.915 & 0.644 & 86.3\% & 0.933 & 0.853 & 0.724 \\
EfficientNetB5 & 71.0\% & 66.2\% & 0.922 & 0.648 & 87.3\% & 0.936 & 0.863 & 0.744 \\
MobileNetV2 & 61.9\% & 60.5\% & 0.894 & 0.572 & 83.3\% & 0.908 & 0.821 & 0.665 \\
ViTBase & 64.9\% & 63.6\% & 0.898 & 0.600 & 83.3\% & 0.909 & 0.822 & 0.665 \\
ViTLarge & 72.6\% & 68.4\% & 0.930 & 0.676 & \underline{87.5\%} & 0.940 & \underline{0.866} & \underline{0.748} \\
DinoV2Small & 69.6\% & 66.4\% & 0.915 & 0.651 & 86.2\% & 0.929 & 0.856 & 0.725 \\
DinoV2Base & \underline{73.8\%} & \underline{71.4\%} & 0.930 & \underline{0.698} & 87.0\% & 0.935 & 0.861 & 0.740 \\
DinoV2Large & \underline{73.8\%} & 68.2\% & \underline{0.933} & 0.678 & 86.6\% & \underline{0.941} & 0.857 & 0.733 \\
\midrule
Ensemble & \textbf{75.6\%} & \textbf{71.9\%} & \textbf{0.937} & \textbf{0.716} & \textbf{88.0\%} & \textbf{0.951} & \textbf{0.880} & \textbf{0.775} \\
\bottomrule
\end{tabular}
}
\vspace{0.15cm}
\end{table*}

\end{appendices}

\end{document}